# Nonparametric Link Prediction in Dynamic Networks

**Purnamrita Sarkar**[*]  PSARKAR@EECS.BERKELEY.EDU
**Deepayan Chakrabarti**[†]  DEEPAY@FB.COM
**Michael I. Jordan**[*,‡]  JORDAN@EECS.BERKELEY.EDU

[*]Department of EECS and [‡]Department of Statistics, University of California, Berkeley
[†]Facebook (This work was partly done when the author was at Yahoo! Research)

## Abstract

We propose a nonparametric link prediction algorithm for a sequence of graph snapshots over time. The model predicts links based on the features of its endpoints, as well as those of the *local neighborhood* around the endpoints. This allows for different *types* of neighborhoods in a graph, each with its own dynamics (e.g, growing or shrinking communities). We prove the consistency of our estimator, and give a fast implementation based on locality-sensitive hashing. Experiments with simulated as well as five real-world dynamic graphs show that we outperform the state of the art, especially when sharp fluctuations or nonlinearities are present.

## 1. Introduction

The problem of predicting links in a graph occurs in many settings—recommending friends in social networks, predicting movies or songs to users, market analysis, and so on. However, state-of-the-art methods suffer from two weaknesses. First, most methods rely on heuristics such as counting common neighbors, etc. while these often work well in practice, their theoretical properties have not been thoroughly analyzed. (Sarkar et al. (2010) is one step in this direction). Second, most of the heuristics are meant for predicting links from one static snapshot of the graph. However, graph datasets often carry additional temporal information such as the creation and deletion times of nodes and edges, so the data is better viewed as a sequence of snapshots of an evolving graph or as a continuous time process (Vu et al., 2011). In this paper, we focus on link prediction in the sequential snapshot setting, and propose a nonparametric method that (a) makes weak model assumptions about the graph generation process, (b) leads to formal guarantees of consistency, and (c) offers a fast and scalable implementation via locality sensitive hashing (LSH).

Our approach falls under the framework of nonparametric time series prediction, which models the evolution of a sequence $x_t$ over time (Masry & Tjøstheim, 1995). Each $x_t$ is modeled as a function of a moving window $(x_{t-1}, \ldots, x_{t-p})$, and so $x_t$ is assumed to be independent of the rest of the time series given this window; the function itself is learned via kernel regression. In our case, however, there is a graph snapshot in each timestep. The obvious extension of modeling each graph as a multi-dimensional $x_t$ quickly runs into problems of high dimensionality, and is not scalable. Instead, we appeal to the following intuition: the graphs can be thought of as providing a "spatial" dimension that is orthogonal to the time axis. In the spirit of the time series model discussed above, our model makes the additional assumption that the linkage behavior of any node $i$ is independent of the rest of the graph given its "local" neighborhood or cluster $N(i)$; in effect, local neighborhoods are to the spatial dimension what moving windows are to the time dimension. Thus, the out-edges of $i$ at time $t$ are modeled as a function of the local neighborhood of $i$ over a moving window, resulting in a much more tractable problem. This model also allows for different *types* of neighborhoods to exist in the same graph, e.g., regions of slow versus fast change in links, assortative versus disassortative regions (where high-degree nodes are more/less likely to connect to other high-degree nodes), densifying versus sparsifying regions, and so on. An additional advantage of our nonparametric model is that it can easily incorporate node and link features which are not based on the graph topology (e.g., labels in labeled graphs).

Our contributions are as follows:
(1) *Nonparametric problem formulation:* We offer, to our knowledge, the first nonparametric model for link prediction in dynamic graphs. The model is powerful enough to accommodate regions with very different evolution profiles, which would be impossible for any single link prediction rule or heuristic. It also enables learning based on both topological as well as other externally available features (such as labels).

(2) *Consistency of the estimator:* Using arguments from the literature on Markov chains and strong mixing, we prove consistency of our estimator.

(3) *Fast implementation via LSH:* Nonparametric methods such as kernel regression can be very slow when the kernel must be computed between a query and all points in the training set. We adapt the locality sensitive hashing algorithm of Indyk & Motwani (1998) for our particular kernel function, which allows the link prediction algorithm to scale to large graphs and long sequences.

(4) *Empirical improvements over previous methods:* We show that on graphs with nonlinearities, such as seasonally fluctuating linkage patterns, we outperform all of the state-of-the-art heuristic measures for static and dynamic graphs. This result is confirmed on a real-world sensor network graph as well as via simulations. On other real-world datasets whose evolution is far smoother and simpler, we perform as well as the best competitor. Finally, on simulated datasets, our LSH-based kernel regression is shown to be much faster than the exact version while yielding almost identical accuracy. For larger real-world datasets, the exact kernel regression did not even finish in a day.

The rest of the paper is organized as follows. We present the model and prove consistency in Sections 2 and 3. We discuss our LSH implementation in Section 4. We give empirical results in Section 5, followed by related work and conclusions in Sections 6 and 7.

## 2. Proposed Method

Consider the link prediction problem in static graphs. Simple heuristics like picking node pairs that were linked most recently (i.e., had small time to last-link), or that have the most common neighbors, have been shown empirically to be good indicators of future links between node pairs (Liben-Nowell & Kleinberg, 2003). An obvious extension to dynamic graphs is to compute the fraction of pairs that had lastlink $= k$ at time $t$ and formed an edge at time $t + 1$, aggregated over all timesteps $t$, and use the value of $k$ with the highest fraction as the best predictor. This can easily be extended to multiple features. Thus, *modulo fraction estimation errors*, the dynamic link prediction problem reduces to the computation and analysis of multi-dimensional histograms, or *datacubes*.

However, this simple solution suffers from two critical problems. First, it does not allow for local variations in the link-formation fractions. This can be addressed by computing a separate datacube for each *local neighborhood* (made more precise later). The second, more subtle, problem is that the above method implicitly assumes stationarity, i.e., a node's link-formation probabilities are *time-invariant* functions of the datacube features. This is clearly inaccurate: it does not allow for seasonal changes in linkage patterns, or for a transition from slow to fast evolution, etc. The solution is to use the datacubes not to directly predict future links, but as a *signature* of the recent evolution of the neighborhood. We can then find historical neighborhoods from some previous time $t$ that had the same signature, and use their evolution from $t$ to $t+1$ to predict link formation in the next timestep for the current neighborhood. Thus, seasonalities and other arbitrary patterns can be learned. Also, this combats sparsity by aggregating data across similarly-evolving communities even if they are separated by graph distance and time. Finally, note that the signature encodes the recent evolution of a neighborhood, and not just the distribution of features in it. Thus, it is *evolution* that drives the estimation of linkage probabilities.

We now formalize these ideas. Let the observed sequence of directed graphs be $\mathcal{G} = \{G_1, G_2, \ldots, G_t\}$. Let $Y_t(i,j) = 1$ if the edge $i \to j$ exists at time $t$, and let $Y_t(i,j) = 0$ otherwise. Let $N_t(i)$ be the local neighborhood of node $i$ in $G_t$; in our experiments, we define it to be the set of nodes within 2 hops of $i$, and all edges between them. Note that the neighborhoods of nearby nodes can overlap. Let $\vec{N}_{t,p}(i) = \{N_t(i), \ldots, N_{t-p+1}(i)\}$. Then, our model is:

$$Y_{t+1}(i,j)|\mathcal{G} \sim \text{Bernoulli}(g(\psi_t(i,j)))$$
$$\psi_t(i,j) = \{s_t(i,j), d_t(i)\},$$

where $0 \leq g(.) \leq 1$ is a function of two sets of features: those specific to the *pair* of nodes $(i,j)$ under consideration ($s_t(i,j)$), and those for the *local neighborhood* of the endpoint $i$ ($d_t(i)$). We require that both of these be functions of $\vec{N}_{t,p}(i)$. Thus, $Y_{t+1}(i,j)$ is assumed to be independent of $\mathcal{G}$ given $\vec{N}_{t,p}(i)$, limiting the dimensionality of the problem. Also, two pairs of nodes $(i,j)$ and $(i',j')$ that are close to each other in terms of graph distance are likely to have overlapping neighborhoods, and hence higher chances of sharing neighborhood-specific features. Thus, link prediction probabilities for pairs of nodes from the same graph region are likely to be dependent, as expected.

Assume that the pair-specific features $s_t(i,j)$ come from a finite set $S$; if not, they are discretized into such a set. For example, one may use $s_t(i,j) = \{\text{cn}_t(i,j), \ell\ell_t(i,j)\}$ (i.e., number of common neighbors and the last time a link appeared between nodes $i$ and $j$). Let $d_t(i) = \{\eta_{it}(s), \eta_{it}^+(s) \, \forall s \in S\}$, where $\eta_{it}(s)$ are the number of node pairs in $N_{t-1}(i)$ with feature vector $s$, and $\eta_{it}^+(s)$ the number of such pairs which were also linked by an edge in the next timestep $t$. In a nutshell, $d_t(i)$ tells us the chances of an edge being created in $t$ given its features in $t-1$, averaged over the whole neighborhood $N_{t-1}(i)$ — in other words, it captures the *evolution* of the neighborhood around $i$ over one timestep.

One can think of $d_t(i)$ as a multi-dimensional histogram, or a "datacube", which is indexed by the features $s$. Hence, now onwards we will often refer to $d_t(i)$ as a "datacube", and a feature vector $s$ as the "cell" $s$ in the datacube with contents $(\eta_{it}(s), \eta_{it}^+(s))$. Finiteness of $S$ is necessary to ensure that datacubes are finite-dimensional, which allows us to index them and quickly find nearest-neighbor datacubes.

ESTIMATOR. Our estimator of the function $g(.)$ is:

$$\hat{g}(\psi_t(i,j)) = \frac{\sum_{i',j',t'} \text{Sim}(\psi_t(i,j), \psi_{t'}(i',j')) \cdot Y_{t'+1}(i',j')}{\sum_{i',j',t'} \text{Sim}(\psi_t(i,j), \psi_{t'}(i',j'))}.$$

To reduce dimensionality, we factor $\text{Sim}(\psi_t(i,j), \psi_{t'}(i',j'))$ into neighborhood-specific and pair-specific parts: $K(d_t(i), d_{t'}(i')) \cdot I\{s_t(i,j) = s_{t'}(i',j')\}$. In other words, the similarity measure $\text{Sim}(.)$ computes the similarity between the two neighborhood evolutions (i.e., the datacubes), but only for pairs $(i',j')$ at time $t'$ that had exactly the same features as the query pair $(i,j)$ at $t$ (i.e., pairs belonging to the cell $s = s_t(i,j)$). This yields a different interpretation of the estimator:

$$\frac{\sum_{i',t'} K(d_t(i), d_{t'}(i')) \cdot \sum_{j'} [I\{s_t(i,j) = s_{t'}(i',j')\} \cdot Y_{t'+1}(i',j')]}{\sum_{i',t'} K(d_t(i), d_{t'}(i')) \cdot \sum_{j'} I\{s_t(i,j) = s_{t'}(i',j')\}}$$

$$= \frac{\sum_{i',t'} K(d_t(i), d_{t'}(i')) \cdot \eta_{i't'+1}^+(s_t(i,j))}{\sum_{i',t'} K(d_t(i), d_{t'}(i')) \cdot \eta_{i't'+1}(s_t(i,j))}.$$

Intuitively, given the query pair $(i,j)$ at time $t$, we look only inside cells for the query feature $s = s_t(i,j)$ in all neighborhood datacubes, compute the average $\eta_{i't'}^+(s)$ and $\eta_{i't'}(s)$ in these cells after accounting for the similarities of the datacubes to the query neighborhood datacube, and use their quotient as the estimate of linkage probability. Thus, the probabilities are computed from historical instances where (a) the feature vector of the historical node pair matches the query, and (b) the local neighborhood was similar as well.

Now, we need a measure of the closeness between neighborhoods. Two neighborhoods are close if they have similar probabilities $p(s)$ of generating links between node pairs with feature vector $s$, for any $s \in S$. We could simply compare point estimates $p(s) = \eta_.^+(s)/\eta_.(s)$, but this does not account for the variance in these estimates. Instead, we consider the full posterior of $p(s)$ (a Beta distribution), and use the total variation distance between these Betas as a measure of the closeness:

$$K(d_t(i), d_{t'}(i')) = b^{D(d_t(i), d_{t'}(i'))} \quad (0 < b < 1) \quad (1)$$
$$D(d_t(i), d_{t'}(i')) = \sum_{s \in S} \text{TV}(X, Y)$$
$$X \sim \mathcal{B}(\eta_{it}^+(s), \eta_{it}(s) - \eta_{it}^+(s))$$
$$Y \sim \mathcal{B}(\eta_{i't'}^+(s), \eta_{i't'}(s) - \eta_{i't'}^+(s)),$$

where $\text{TV}(.,.)$ denotes the total variation distance between the distributions of its two argument random variables, and $b \in (0,1)$ is a bandwidth parameter.

DEALING WITH SPARSITY. For sparse graphs, or short time series, two practical problems can arise. First, a node $i$ could have zero degree and hence an empty neighborhood. In order to get around this, we define the neighborhood of node $i$ as the union of 2-hop neighborhoods over the last $p$ timesteps.

Second, the $\eta_.(s)$ and $\eta_.^+(s)$ values obtained from kernel regression could be too small, and so the estimated linkage probability $\eta_.^+(s)/\eta_.(s)$ is too unreliable for prediction and ranking. We offer a threefold solution. (a) We combine $\eta_.(s)$ and $\eta_.^+(s)$ with a weighted average of the corresponding values for any $s'$ that are "close" to $s$, the weights encoding the similarity between $s'$ and $s$. This is in essence the same as replacing the indicator in Eq. (1) with a kernel that measures similarity between features. (b) Instead of ranking node pairs using $\eta_.^+(s)/\eta_.(s)$, we use the lower end of the 95% Wilson score interval (Wilson, 1927), which is a widely used *binomial proportion confidence interval*. The node pairs that are ranked highest according to this "Wilson score" are those that have high estimated linkage probability $\eta_.^+(s)/\eta_.(s)$ *and* $\eta_.(s)$ is high (implying a reliable estimate). (c) Last but not the least, we maintain a "prior" datacube, which is average of all historical datacubes. The Wilson score of each node pair is smoothed with the corresponding score derived from the prior datacube, with the degree of smoothing depending on $\eta_.(s)$. This can be thought of as a simple hierarchical model, where the lower level (set of individual datacubes) smooths its estimates using the higher level (the prior datacube).

## 3. Consistency of Kernel Estimator

Now, we prove that the estimator $\widehat{g}$ defined in Eq. (1) is consistent. Recall that our model is as follows:

$$Y_{t+1}(i,j)|\mathcal{G} \sim \text{Ber}(g(\psi_t(i,j))), \quad (2)$$

where $\psi_t(i,j)$ equals $\{s_t(i,j), d_t(i)\}$. Assume that all graphs have $n$ nodes ($n$ is finite). Let $Q$ represent the query datacube $d_T(q)$. We want to obtain predictions for timestep $T+1$. From Eq. (1), the kernel estimator of $g$ for query pair $(q, q')$ at time $T+1$ can be written as:

$$\hat{g}(s, Q) = \frac{\hat{h}(s, Q)}{\hat{f}(s, Q)} \quad \text{(where } s = s_T(q, q'))$$

$$\hat{h}(s, Q) = \frac{1}{n(T-p)} \sum_{t=p}^{T-1} \sum_{i=1}^{n} K_b(d_t(i), Q) \eta_{it+1}^+(s)$$

$$\hat{f}(s, Q) = \frac{1}{n(T-p)} \sum_{t=p}^{T-1} \sum_{i=1}^{n} K_b(d_t(i), Q) \eta_{it+1}(s).$$

The estimator $\hat{g}$ is defined only when $\hat{f} > 0$. The kernel was defined earlier as $K_b(d_t(i), Q) = b^{D(d_t(i), Q)}$, where the bandwidth $b$ tends to 0 as $T \to \infty$, and $D(.)$ is the distance function defined in Eq. (1). This is similar to other discrete kernels (Aitchison & Aitken, 1976), and has the following property

$$\lim_{b \to 0} K_b(d_t(i), Q) = \begin{cases} 1 & \text{if } d_t(i) = Q \\ 0 & \text{otherwise.} \end{cases} \quad (3)$$

**Theorem 3.1** (Consistency). *$\hat{g}$ is a consistent estimator of $g$, i.e., $\hat{g} \xrightarrow{P} g$ as $T \to \infty$.*

*Proof.* The proof is in two parts. Lemma 3.3 will show that $\text{var}(\hat{h})$ and $\text{var}(\hat{f})$ tend to 0 with $T \to \infty$. Lemma 3.4 shows that their expectations converge to $g(s, Q)R$ and $R$ respectively, for some constant $R > 0$. Hence, $(\hat{h}, \hat{f}) \xrightarrow{P} (g(s, Q)R, R)$. By the continuous mapping theorem, $\hat{g} = \hat{h}/\hat{f} \xrightarrow{P} g$. □

The next lemma upper bounds the growth of variance terms. We first recall the concept of strong mixing. For a Markov chain $S_t$, define the strong mixing coefficients $\alpha(k) \doteq \sup_{|t-t'| \geq k} \{|P(A \cap B) - P(A)P(B)| : A \in \mathcal{F}_{\leq t}, B \in \mathcal{F}_{\geq t'}\}$, where $\mathcal{F}_{\leq t}$ and $\mathcal{F}_{\geq t'}$ are the sigma algebras generated by events in $\bigcup_{i \leq t} S_i'$ and $\bigcup_{i \geq t'} S_i'$ respectively. Intuitively, small values of $\alpha(k)$ imply that states that are $k$ apart in the Markov chain are almost independent. For bounded $A$ and $B$, this also limits their covariance: $|\text{cov}(A, B)| \leq c\alpha(k)$ for some constant $c$ (Durrett, 1995).

**Lemma 3.2.** *Let $q_{it}$ be a bounded function of $\eta_{it+1}(s), \eta_{it+1}^+(s)$ and $d_t(i)$. Then, $(1/T^2) \text{var}\left[\sum_{t=1}^{T} \sum_{i=1}^{n} q_{it}\right] \to 0$ as $T \to \infty$.*

*Proof Sketch.* Our graph evolution model is Markovian; assuming each "state" to represent the past $p+1$ graphs, the next graph (and hence the next state) is a function only of the current state. The state space is also finite, since each graph has bounded size. Thus, the state space may be partitioned as $\mathcal{S} = TR \bigcup C_i$, where $TR$ is a set of transient states, each $C_i$ is an irreducible closed communication class, and there exists at least one $C_i$ (Grimmett & Stirzaker, 2001).

The Markov chain must eventually enter some $C_i$. First assume that this class is aperiodic. Irreducibility and aperiodicity implies geometric ergodicity (Fill, 1991), which implies strong mixing with exponential decay (Pham, 1986): $\alpha(k) \sim e^{-\beta k}$ for some $\beta > 0$. Thus, $\sum_{t,t'} \text{cov}(q_{it}, q_{jt'}) \approx \sum_t \sum_{|t-t'|=0}^{T} \text{cov}(q_{it}, q_{jt'}) \leq \sum_t \sum_{k=0}^{\infty} c\alpha(k) = \sum_t \sum_k ce^{-\beta k} = O(T)$. Thus, $\text{var}(\sum_t \sum_i q_{it})/T^2 = O(1/T)$, which goes to zero as $T \to \infty$. The proof for a cyclic communication class, while similar in principle, is more involved and is deferred to the appendix. □

**Lemma 3.3.** *$\text{var}(\hat{h})$ and $\text{var}(\hat{f})$ tend to 0 as $T \to \infty$.*

*Proof.* The result follows by applying Lemma 3.2 with $q(.)$ equal to $K_b(d_t(i), Q)\eta_{it+1}^+(s)$ and $K_b(d_t(i), Q)\eta_{it+1}(s)$ respectively. □

**Lemma 3.4.** *As $T \to \infty$, for some $R > 0$,*
$$E[\hat{h}(s, Q)] \to g(s, Q)R, \quad E[\hat{f}(s, Q)] \to R.$$

*Proof.* Let $\epsilon$ denote the minimum distance between two datacubes that are not identical; since the set of all possible datacubes is finite, $\epsilon > 0$. $E[\hat{h}(s, Q)]$ is an average of terms $E[K_b(d_t(i), Q)\eta_{it+1}^+(s)]$, over $i \in \{1, \ldots, n\}$ and $t \in \{p, \ldots, T-1\}$. Now, $E[K_b(d_t(i), Q)\eta_{it+1}^+(s)] = E\left[b^{D(d_t(i), Q)} E\left[\eta_{it+1}^+(s) | d_t(i)\right]\right]$, and the inner expectation is $E[\eta_{it+1}^+(s) \cdot g(s, d_t(i))]$, as can be seen by summing Eq. (2) over all pairs $(i, j)$ in a neighborhood with identical $s_t(i, j)$, and then taking expectations. Writing the expectation in terms of a sum over all possible datacubes, and noting that everything is bounded, gives the following:

$$E\left[b^{D(d_t(i), Q)} \eta_{it+1}(s) \cdot g(s, d_t(i))\right]$$
$$= E[\eta_{it+1}(s) | d_t(i) = Q] \cdot g(s, Q) P(d_t(i) = Q) + O(b^{\epsilon}).$$

Recalling that $E[\hat{h}(s, Q)]$ was an average of the above terms, $E[\hat{h}(s, Q)]$ equals the following.

$$\frac{g(s, Q) \sum_{t,i} E[\eta_{it+1}(s) | d_t(i) = Q] \cdot P(d_t(i) = Q)}{n(T-p)} + O(b^{\epsilon}).$$

Using the argument of Lemma 3.2, we will eventually hit a closed communication class. Also, the query datacube at $T$ is a function of the state $S_T$, which belongs to a closed irreducible set $C$ with probability 1. Hence, using standard properties of finite state space Markov chains (in particular positive recurrence of states in $C$), we can show that the above average converges to a positive constant $R$ times $g(s, Q)$. An identical argument yields $E[\widehat{f}(s,Q)]$ converges to $R$. The full proof can be found in the appendix. □

## 4. Fast search using LSH

A naive implementation of the nonparametric estimator in Eq. (1) searches over all $n$ datacubes for each of the $T$ timesteps for each prediction, which can be very slow for large graphs. In most practical situations, the top-$r$ closest neighborhoods should suffice (in our case $r = 20$). Thus, we need a fast sublinear-time method to quickly find the top-$r$ closest neighborhoods.

We achieve this via locality sensitive hashing (LSH) (Indyk & Motwani, 1998). The standard LSH operates on bit sequences, and maps sequences with small Hamming distance to the same hash bucket. However, we must hash datacubes, and use the total variation distance metric. Our solution is based on the fact that total variation distance between discrete distributions is half the $L_1$ distance between the corresponding probability mass functions. If we could approximate the probability distributions in each cell with bit sequences, then the $L_1$ distance would just be the Hamming distance between these sequences, and standard LSH could be used for our datacubes.

CONVERSION TO BIT SEQUENCE. The key idea is to approximate the linkage probability distribution by its histogram. We first discretize the range $[0,1]$ (since we deal with probabilities) into $B_1$ buckets. For each bucket we compute the probability mass $p$ falling inside it. This $p$ is encoded using $B_2$ bits by setting the first $\lfloor pB_2 \rfloor$ bits to 1, and the others to 0. Thus, the entire distribution (i.e., one cell) is represented by $B_1B_2$ bits. The entire datacube can be stored in $|S|B_1B_2$ bits. However, in all our experiments, datacubes were very sparse with only $M \ll |S|$ cells ever being non-empty (usually, 10-50); thus, we use only $MB_1B_2$ bits in practice. The Hamming distance between two pairs of $MB_1B_2$ bit vectors yields the total variation distance between datacubes (modulo a constant factor).

DISTANCES VIA LSH. We create a hash function by just picking a uniformly random sample of $k$ bits out of $MB_1B_2$. For each hash function, we create a hash table that stores all datacubes whose hashes are identical in these $k$ bits. We use $\ell$ such hash functions. Given a query datacube, we hash it using each of these $\ell$ functions, and then create a *candidate set* of up to $O(\max(\ell, r))$ of distinct datacubes that share any of these $\ell$ hashes. The total variation distance of these candidates to the query datacube is computed explicitly, yielding the closest matching historical datacubes.

PICKING $k$. The number of bits $k$ is crucial in balancing accuracy versus query time: a large $k$ sends all datacubes to their own hash bucket, so any query can find only a few matches, while a small $k$ bunches many datacubes into the same bucket, forcing costly and unnecessary computations of the exact total variation distance. We do a binary search to find the $k$ for which the average hash-bucket size over a query workload is just enough to provide the desired top-20 matches. Its accuracy is shown in Section 5.

Finally, we underscore a few points. First, the entire bit representation of $MB_1B_2$ bits never needs to be created explicitly; only the hashes need to be computed, and this takes $O(k\ell)$ time. Second, the main cost in the algorithm is in creating the hash table, which needs to be done once as a preprocessing step. Query processing is extremely fast and sublinear, since the candidate set is much smaller than the size of the training set. Finally, we have found the loss due to approximation to be minimal in all our experiments.

## 5. Experiments

We first introduce several baseline algorithms, and the evaluation metric. We then show via simulations that our algorithm outperforms prior work in a variety of situations modeling nonlinearities in linkage patterns, such as seasonality in link formation. These findings are confirmed on several evolving real-world graphs: a sensor network, two co-authorship graphs, and a stock return correlation graph. Finally, we demonstrate the improvement in timing achieved via LSH over exact search, and the effect of LSH bit-size $k$ on accuracy.

BASELINES AND METRICS. We compare our nonparametric link prediction algorithm (NonParam) to the following baselines which, though seemingly simple, are extremely hard to beat in practice (Liben-Nowell & Kleinberg, 2003; Tylenda et al., 2009):

LL: ranks pairs using ascending order of *last time of linkage* (Tylenda et al., 2009).

CN (last timestep): ranks pairs using descending order of the number of *common neighbors* (Liben-Nowell & Kleinberg, 2003).

AA (last timestep): ranks pairs using descending order of the *Adamic-Adar* score (Adamic & Adar, 2003), a weighted variant of common neighbors which it has been shown to outperform (Liben-Nowell & Kleinberg, 2003).

Katz (last timestep): extends CN to paths with length greater than two, but with longer paths getting exponentially smaller weights (Katz, 1953).

CN-all, AA-all, Katz-all: CN, AA, and Katz computed on *the union of all graphs until the last timestep*.

Recall that, for NonParam, we only predict on pairs which are in the neighborhood (generated by the union of 2-hop neighborhoods of last $p$ timesteps) of each other. We deliberately used a simple feature set for NonParam, setting $s_t(i,j) = \{cn_t(i,j), \ell\ell_t(i,j)\}$ (i.e., common neighbors and last-link) and not using any external "meta-data" (e.g., stock sectors, university affiliations, etc.). All feature values are binned logarithmically in order to combat sparsity in the tails of the feature distributions. Mathematically, our feature $\ell_t(i,j)$ should be capped at $p$. However, since the common heuristic LL uses no such capping, for fairness, we used the uncapped 'last time a link appeared' as $\ell_t(i,j)$, for the pairs we predict on. The bandwidth $b$ is picked by cross-validation.

For any graph sequence $(G_1,\ldots,G_T)$, we test link prediction accuracy on $G_T$ for a subset $S_{>0}$ of nodes with non-zero degree in $G_T$. Each algorithm is provided training data until timestep $T-1$, and must output, for each node $i \in S_{>0}$, a ranked list of nodes in descending order of probability of linking with $i$ in $G_T$. For purposes of efficiency, we only require a ranking on the nodes that have ever been within two hops of $i$ (call these the candidate pairs); all algorithms under consideration predict the absence of a link for nodes outside this subset. We compute the AUC score for predicted scores for all candidate pairs against their actual edges formed in $G_T$.

### 5.1. Prediction Accuracy

We compare accuracy on (a) simulations, (b) a real-world sensor network with periodicities, and (c) broadly stationary real-world graphs.

SIMULATIONS. One unique aspect of NonParam is its ability to predict even in the presence of sharp fluctuations. As an example, we focus on seasonal patterns, simulating a model of Hoff (personal communication) that posits an independently drawn "feature vector" for each node. Time moves over a repeating sequence of seasons, with a different set of features being "active" in each. Nodes with these features are more likely to be linked in that season, though noisy links also exist. The user features also change smoothly over time, to reflect changing user preferences.

We generated 100-node graphs over 20 timesteps using 3 seasons, and plotted AUC averaged over 10 random runs for several noise-to-signal ratios (Fig. 1). NonParam consistently outperforms all other baselines by a large margin. Clearly, seasonal graphs have non-linear linkage patterns: the best predictor of links at time $T$ are the links at times $T-3$, $T-6$, etc., and NonParam is able to automatically learn this pattern. However, CN, AA, Katz are biased towards predicting links between pairs which were linked (or had short paths connecting them) at the previous timestep $T-1$; this implicit smoothness assumption makes them suffer heavily. This is why they behave as bad as a random predictor (AUC 0.5).

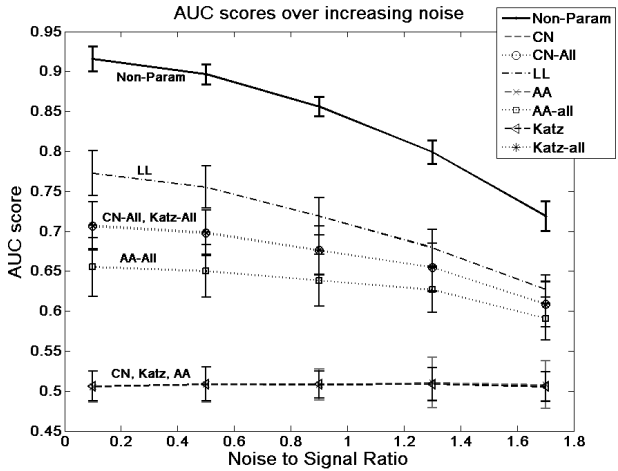

Figure 1. Simulated graphs: Effect of noise

Baselines LL, CN-all, AA-all and Katz-all use information from the union of all graphs until time $T-1$. Since the off-seasonal noise edges are not sufficiently large to form communities, most of the new edges come from communities of nodes created in season. This is why CN-all, AA-all and Katz-all outperform their 'last-timestep' counterparts. As for LL, since links are more likely to come from the last seasons, it performs well, although poorly compared to NonParam. Also note that the changing user features forces the community structures to change slowly over time; in our experiments, CN-all performs worse that it would when there was no change in the user features, since the communities stayed the same.

Table 1 compares average AUC scores for graphs with and without seasonality, using the lowest noise setting from Fig. 1. As already mentioned, CN, AA, Katz perform very poorly on the seasonal graphs, because of their implicit assumption of smoothness. Their variants CN-all, AA-all and Katz-all on the other hand take into account all the community structures seen in the data until the last timestep, and hence are better. On the other hand, for Stationary, links formed in the last few timesteps of the training data are good predictors of future links, and so LL, CN, AA and Katz all perform extremely well. Interestingly, CN-all, AA-all and Katz-all are worse than their 'last time-step' variants owing to the slow movement of the user features. We note, however, that NonParam performs very well in all cases, the margin of improvement being most for the seasonal networks.

|           | Seasonal (T=20) | Stationary (T=20) |
|-----------|-----------------|-------------------|
| NonParam  | **.91 ± .01**   | **0.99 ± .005**   |
| LL        | .77 ± .03       | 0.97 ± .006       |
| CN        | .51 ± .02       | 0.97 ± .01        |
| AA        | .51 ± .02       | 0.95 ± .02        |
| Katz      | .50 ± .02       | 0.97 ± .01        |
| CN-all    | .71 ± .03       | 0.86 ± .03        |
| AA-all    | .65 ± .04       | 0.71 ± .04        |
| Katz-all  | .71 ± .03       | 0.87 ± .03        |

Table 1. Avg. AUC with and without seasonality.

|           | NIPS | HepTh | Citeseer | S&P500 |
|-----------|------|-------|----------|--------|
| NonParam  | **.87** | **.89** | .89    | .73    |
| LL        | .84  | .87   | **.90**  | .70    |
| CN        | .74  | .76   | .69      | .72    |
| AA        | .84  | .87   | **.90**  | .70    |
| Katz      | .75  | .83   | .83      | .76    |
| CN-all    | .56  | .62   | .70      | **.79** |
| AA-all    | .77  | .83   | .83      | .76    |
| Katz-all  | .67  | .71   | .81      | **.79** |

Table 2. Avg. AUC in real world **Stationary** graphs

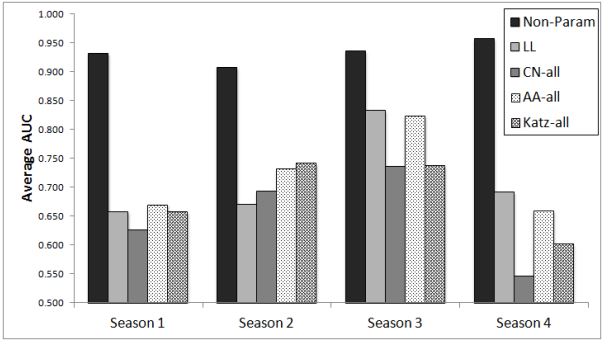

Figure 2. AUC scores for a **periodic** sensor network

REAL-WORLD GRAPHS. We first present results on a 24 node sensor network where each edge represents the successful transmission of a message [1]. We look at up to 82 consecutive measurements. These networks exhibit clear periodicity; in particular, a different set of sensors turn on and communicate during different periods (as our earlier "seasons"). Fig. 5.1 shows our results for these four seasons averaged over several cycles. The maximum standard deviation, averaged over these seasons is .07. We do not show CN, AA and Katz which perform like a random predictor. Non-Param again significantly outperforms the baselines, confirming the simulation results.

Additional experiments were performed on three evolving co-authorship graphs: the Physics "HepTh" community ($n = 14,737$ nodes, $e = 31,189$ total edges, and $T = 8$ timesteps), NIPS ($n = 2,865$, $e = 5,247$, $T = 9$), and authors of papers on Citeseer ($n = 20,912$, $e = 45,672$, $T = 11$) with "machine learning" in their abstracts. Each timestep looks at $1 - 2$ years of papers (so that the median degree at any timestep is at least 1). We also considered an evolving stock-correlation network: the nodes are a subset of stocks in the S&P500, and two stocks are linked if the correlation of their daily returns over a two month window exceeds 0.8 ($n = 424$, $e = 41,699$, $T = 49$).

[1] http://www.select.cs.cmu.edu/data

Table 2 shows the average AUC for all the algorithms. In the co-authorship graphs most authors keep working with a similar set of co-authors, which hides seasonal variations, if any. On these graphs we perform as well or better than LL, which has been shown to be the best heuristic by Tylenda et al. (2009). On the other hand, S&P500 is a correlation graph, so it is not surprising that all the common-neighbor or Katz measures perform very well on them. In particular CN-all and AA-all have the best AUC scores. This is primarily because they count paths through edges that exist in different timesteps, which we do not.

Thus, for graphs lacking a clear seasonal trend, LL is the best baseline on co-authorship graphs but not on the correlation graphs, whereas Katz-all works better on correlation graphs, but poorly on co-authorship graphs. NonParam, however, is the best by a large margin in seasonal graphs, and is better or close to the winner in others.

### 5.2. Usefulness of LSH

The query time per datacube using LSH is extremely small: 0.3s for Citeseer, 0.4s for NIPS, 0.6s for HepTh, and 2s for S&P500. Since exact search is intractable in our large-scale real world data, we demonstrate the speedup of LSH over exact search using simulated data. We also show that the hash bitsize $k$ picked adaptively is the largest value that still gives excellent AUC scores. Since larger $k$ translates to fewer entries per hash bucket and hence faster searches, our $k$ yields the fastest runtime performance as well.

EXACT SEARCH VS. LSH. In Fig. 3(a) we plot the time taken to do top-20 nearest neighbor search for a query datacube. We fix the number of nodes at 100, and increase the number of timesteps. As expected, the exact search time increases linearly with the total number of datacubes, whereas LSH searches in nearly constant time. Also, the AUC score of NonParam with LSH is within 0.4% of that of the exact algorithm on average, implying minimal loss of accuracy from LSH.

NUMBER OF BITS IN HASHING. Fig. 3(b) shows the effectiveness of our adaptive scheme to select the num-ber of hash bits (Section 4). For these experiments, we turned off the smoothing based on the prior datacube. As $k$ increases, the accuracy goes down to 50%, as a result of the fact that NonParam fails to find any matches of the query datacube. Our adaptive scheme finds $k \sim 170$, which yields the highest accuracy.

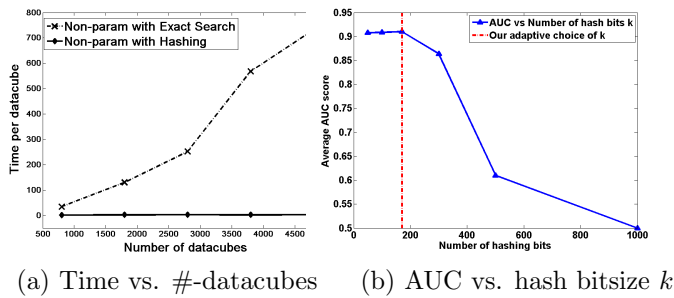

(a) Time vs. #-datacubes    (b) AUC vs. hash bitsize $k$

Figure 3. Time and accuracy using LSH

## 6. Related Work

Existing work on link prediction in dynamic networks can be broadly divided into two categories: generative model based and graph structure based.

GENERATIVE MODELS. These include extensions of Exponential Family Random Graph models (Hanneke & Xing, 2006) by using evolution statistics like edge stability, reciprocity, transitivity; extension of latent space models for static networks by allowing smooth transitions in latent space (Sarkar & Moore, 2005), and extensions of the mixed membership block model to allow a linear Gaussian trend in the model parameters (Fu et al., 2010). In other work, the structure of evolving networks is learned from node attributes changing over time (Kolar et al., 2010). Although these models are intuitive and rich, they generally a) make strong model assumptions, b) require computationally intractable posterior inference, and c) explicitly model linear trends in the network dynamics.

MODELS BASED ON STRUCTURE. Huang & Lin (2009) proposed a linear autoregressive model for individual links, and also built hybrids using static graph similarity features. In Tylenda et al. (2009) the authors examined simple temporal extensions of existing static measures for link prediction in dynamic networks. In both of these works it was shown empirically that LL and its variants are often better or among the best heuristic measures for link prediction. Our nonparametric method has the advantage of presenting a formal model, with consistency guarantees, that also performs as well as LL.

## 7. Conclusions

We proposed a nonparametric model for link prediction in dynamic graphs, and showed that it performs as well as the state of the art for several real-world graphs, and exhibits important advantages over them in the presence of nonlinearities such as seasonality patterns. NonParam also allows us to incorporate features external to graph topology into the link prediction algorithm, and its asymptotic convergence to the true link probability is guaranteed under our fairly general model assumptions. Together, these make NonParam a useful tool for link prediction in dynamic graphs.